\documentclass{article}
\usepackage{ijcai23}

\usepackage{times}
\usepackage{soul}
\usepackage{url}
\usepackage[hidelinks]{hyperref}
\usepackage[utf8]{inputenc}
\usepackage[small]{caption}
\usepackage{multirow}
\usepackage{graphicx}
\usepackage{amsmath}
\usepackage{amsthm}
\usepackage{booktabs}
\usepackage{algorithm}
\usepackage{algorithmic}
\usepackage[switch]{lineno}

\usepackage{color}

\title{From Generation to Suppression: Towards Effective Irregular Glow Removal for Nighttime Visibility Enhancement}

\author{
Wanyu Wu$^1$
\and
Wei Wang$^1${\thanks{Corresponding author: wangwei8@wust.edu.cn}}\and
Zheng Wang$^2$\and
Kui Jiang$^3$\And
Xin Xu$^1$
\affiliations
$^1$Wuhan University of Science and
Technology\\
$^2$Wuhan University\\
$^3$Harbin Institute of Technology
}

\begin{document}

\maketitle
\begin{abstract}
Most existing Low-Light Image Enhancement (LLIE) methods are primarily designed to improve brightness in dark regions, which suffer from severe degradation in nighttime images. However, these methods have limited exploration in another major visibility damage, the \textit{glow effects} in real night scenes.
Glow effects are inevitable in the presence of artificial light sources and cause further diffused blurring when directly enhanced. To settle this issue, we innovatively consider the glow suppression task as \textit{learning physical glow generation} via multiple scattering estimation according to the Atmospheric Point Spread Function (APSF).
In response to the challenges posed by uneven glow intensity and varying source shapes, an APSF-based \textbf{N}ighttime \textbf{I}maging \textbf{M}odel with \textbf{N}ear-field \textbf{L}ight \textbf{S}ources (NIM-NLS) is specifically derived to design a \textit{scalable} \textbf{L}ight-aware \textbf{B}lind \textbf{D}econvolution \textbf{N}etwork (LBDN). The glow-suppressed result is then brightened via a \textbf{R}etinex-based \textbf{E}nhancement \textbf{M}odule (REM).
Remarkably, the proposed glow suppression method is based on zero-shot learning and does not rely on any paired or unpaired training data. 
Empirical evaluations demonstrate the effectiveness of the proposed method in both glow suppression and low-light enhancement tasks.
\end{abstract}
\begin{center} 
\begin{figure}[t]
	\centering
    \footnotesize{
    \tabcolsep=1pt
      \begin{tabular}{cc}
      \includegraphics[width=0.21\textwidth]{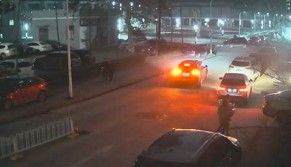} &
      \includegraphics[width=0.21\textwidth]{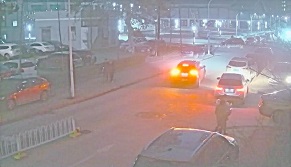} \\
      \small{Input} & \small{DCE~\cite{li2021learning}}\\
      \includegraphics[width=0.21\textwidth]{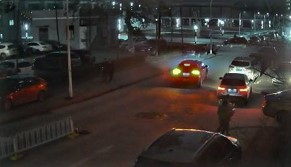} &
      \includegraphics[width=0.21\textwidth]{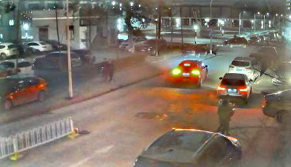} \\
      \small{Our LBDN}& \small{Our REM}\\
      \end{tabular}}
	\caption{Nighttime images suffering from glow effects will be further diffused after existing LLIE methods, significantly impairing the visibility. The proposed LBDN and REM can effectively handle glow effects and boost nighttime visibility.}
\label{methodImg}
\end{figure}
\end{center}
\vspace{-5mm}
\section{Introduction}
Images captured in low light are often accompanied by reduced visibility and information loss in dark areas. Thus, low light image enhancement (LLIE) has sparked a surge of interest. Apart from dim light's impact on image quality, the widespread presence of glow effects around light sources significantly reduces visibility in real-world nighttime scenes.
However, existing LLIE methods devote to low-light areas brightening only and are not well-suited to tackle glow effects.
When applied to images with glow, these methods may actually exacerbate the problem, causing further diffusion of glow areas and introducing artifacts that degrade overall image visibility (Fig.~\ref{methodImg}). To this end, developing effective glow suppression techniques is an urgent task.

Deep learning has proven promising results for LLIE tasks \cite{2020Lightening,jiang2021enlightengan,li2021learning}, while a substantial portion of which incorporates the traditional Retinex model~\cite{jobson1997multiscale} for better luminance boosting through suppression of noise. 
To make a step forward, a large body of work has progressively exerted great effort in more extreme 
conditions~\cite{chen2018learning,zhang2019zero,zhu2020eemefn,wei2020physics}. Regretfully, all these methods cannot produce a satisfactory performance on nighttime images with glows.

In a recent study~\cite{sharma2021nighttime}, the first enhancement method for suppressing glow effects is proposed. This method regards the glow as low-frequency information in the image and adopts a decomposition model ~\cite{wu2018fast} to disentangle the linearized image into low and high-frequency feature maps. In a similar way, the glow in \cite{2207.10564} is removed via high-low frequency decomposition on the assumption of smooth glow in the low-frequency layer.
However, these methods do not consider the fact that glow regions may occur in both low-frequency and high-frequency layers, as is often observed in captured images.

Glows demand a more \textit{universal} description to form a reliable solution. To form a universal definition of the glow suppression task, this study defines the glow based on its physical formation, which arises from a light source scattered several times across medium particles characterized by the atmospheric point spread function (APSF). The first attempt to achieve glow suppression from glow generation is introduced via deep learning-based APSF formation.
\begin{center} 
\begin{figure}[t]
	\centering
    \includegraphics[width=0.44\textwidth]{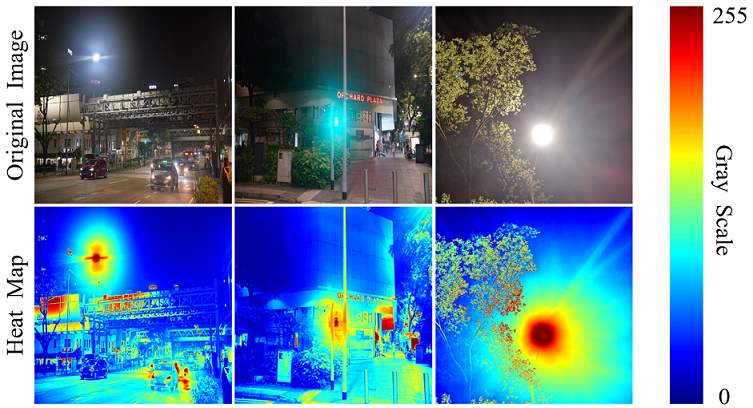} 
	\caption{Some examples suffering from glow effects and their corresponding heat maps. Dark areas and distinct color borders of the glow in the heat map show non-smooth intensity variations.}
\label{heatmap}
\end{figure}
\end{center}
\vspace{-6mm}

Real-world glow effects pose two challenges, the complex variety of light source shapes and the glow intensities unevenly decaying from light sources, as depicted in Fig.~\ref{heatmap}, featuring rough or divergent glows. 
To solve these challenges, our nighttime imaging model with near-field light sources (NIM-NLS) is specifically tailored to account for all multi-scattering lights associated with arbitrary glow effects, direct reflectance from objects, and the impact of ambient light.

Leveraging the physical model, we proceed to solve the problem of recovering both the blurred glow and the clear transmission in NIM-NLS. However, we encountered further challenges, which highlighted limitations in existing methods. Specifically, we observed that: 1) glow only affects local regions around the light source, existing global restoration models cannot recover glow region accurately~\cite{zhou2019kernel,ren2020neural,feng2021removing}; 2) glow patterns are irregular and not sharply bounded, which cannot be extracted by layer separation methods~\cite{levin2007user,zhang2018single,gandelsman2019double}.
Drawing inspiration from these findings, our light-aware blind deconvolution net (LBDN) is designed to construct a local glow estimation. 
Guided by the principle that the center of the glow exhibits maximum intensity, our approach involves initially segmenting the light sources, followed by performing a blind deconvolution on the local glow.
Irregular glow formulation is introduced in our approach with two priors. The light source spatial location mask prior $M_{0}$ indicates spatial locations, while the learning-based APSF prior accommodates for varying glows. Notably, our LBDN can be flexibly extended into existing LLIE methods as a pre-processing module.

\begin{center} 
\begin{figure}[t]
	\centering
    \footnotesize{
    \tabcolsep=1pt
      \begin{tabular}{ccc}
      \includegraphics[width=0.12\textwidth]{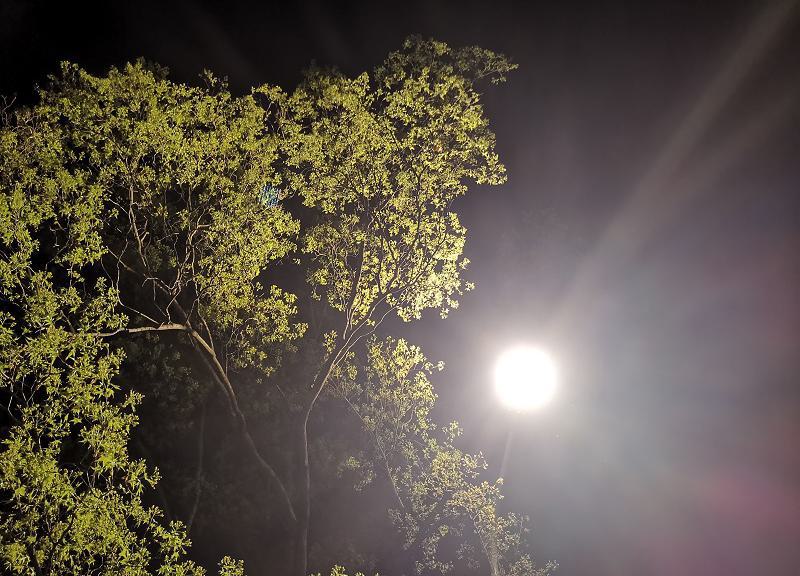} &
      \includegraphics[width=0.12\textwidth]{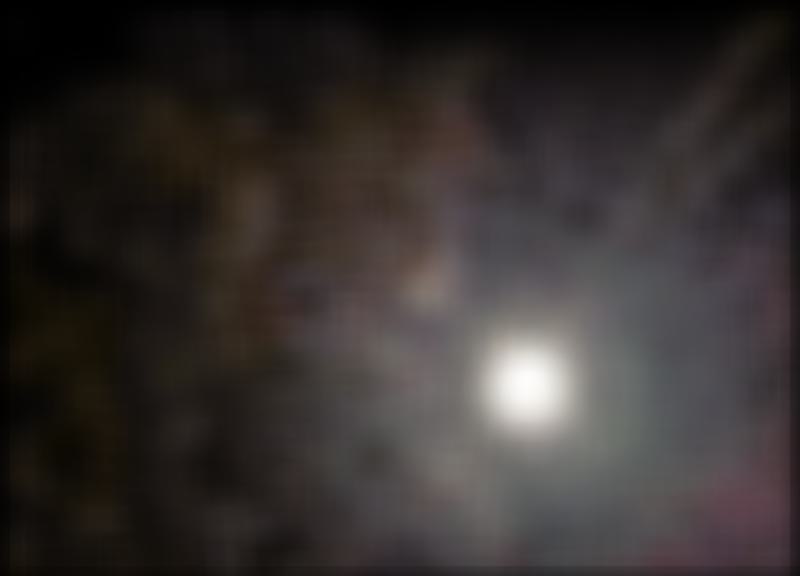} &
      \includegraphics[width=0.12\textwidth]{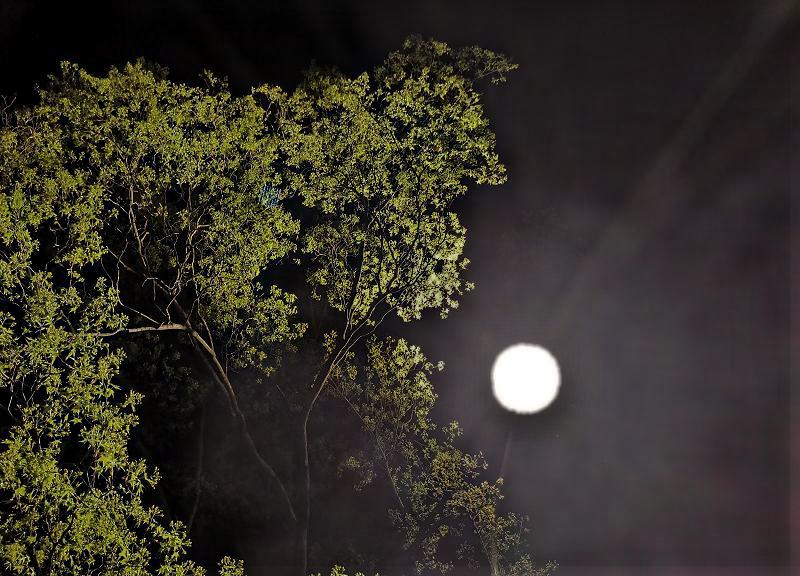} \\
      \includegraphics[width=0.12\textwidth]{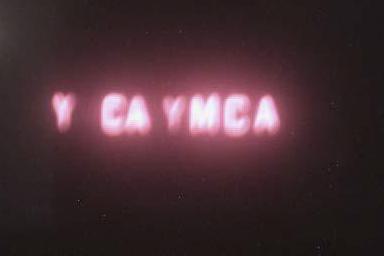} &
      \includegraphics[width=0.12\textwidth]{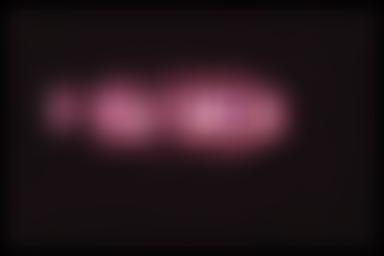} &
      \includegraphics[width=0.12\textwidth]{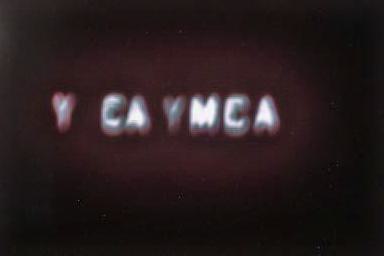} \\
      \includegraphics[width=0.12\textwidth]{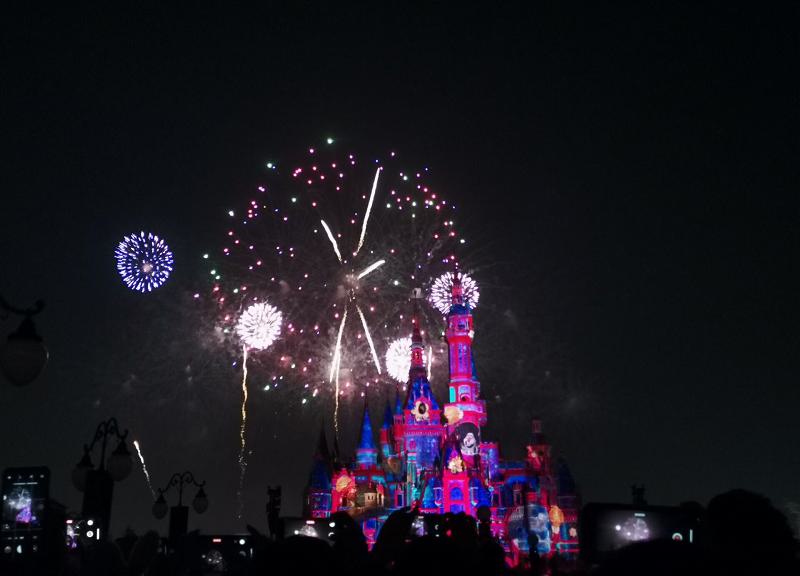} &
      \includegraphics[width=0.12\textwidth]{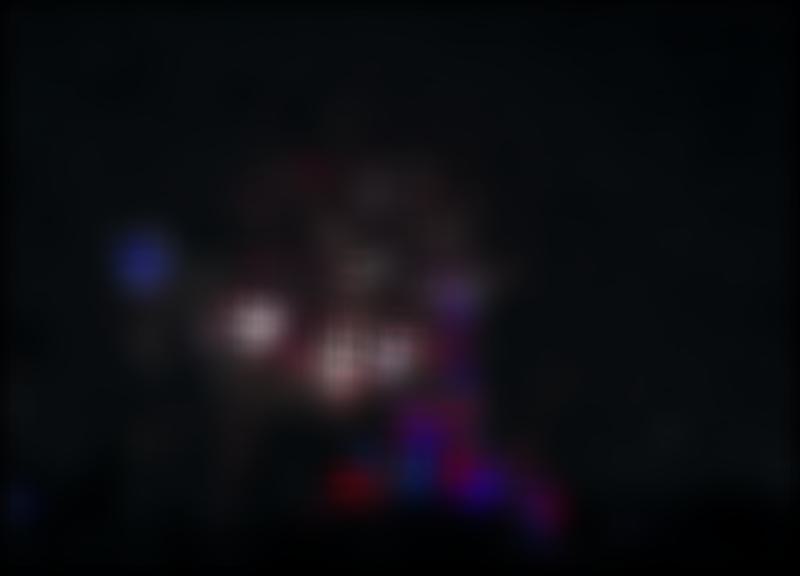} &
      \includegraphics[width=0.12\textwidth]{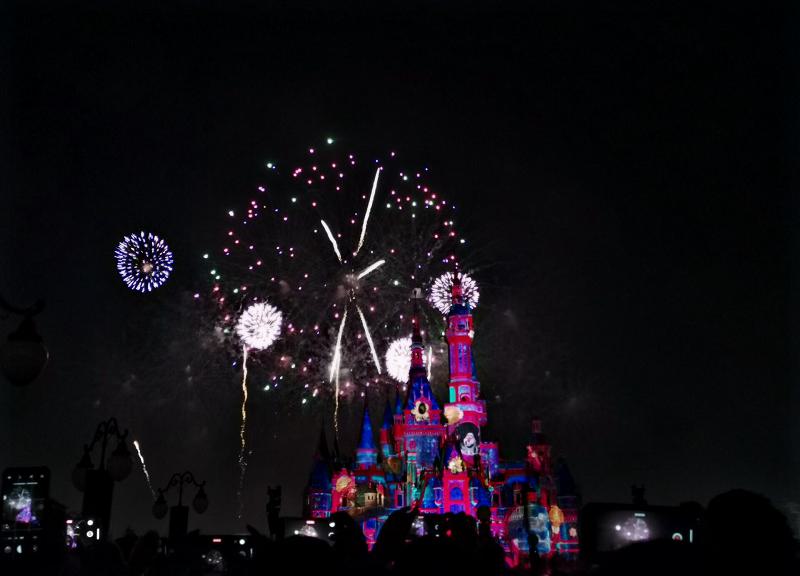} \\
      \footnotesize{Input $I$} & \footnotesize{Multi-scattering $G$}& \footnotesize{Transmission $D$}\\
      \end{tabular}}
	\caption{Examples obtained from our LBDN, demonstrating the effective removal of irregular glow. $G$ is the multi-scattering map estimated by a network, while $D$ is the clear direct transmission map after removing the multi-scattering.}
\label{shape}
\end{figure}
\end{center}

To the best of our knowledge, the proposed method is the first to tackle visibility degradation in various forms of uneven and irregularly shaped glows (see Fig.~\ref{shape}). In the meantime, our approach  is not reliant on paired or unpaired images, nor does it necessitate pre-training. In summary, our contributions can be concluded as:
\begin{itemize}
    \item \textbf{Perspective contribution}. The glow suppression task is innovatively treated as the learning of glow generation, where learning-based APSF formation is utilized.
    \item \textbf{Technical contribution}. Two challenges of arbitrarily shaped light sources and glows with uneven intensity are tackled for the first time. A novel nighttime imaging model with near-field light sources is derived, and solved by a scalable light-aware blind deconvolutional network.
    \item  \textbf{Practical contribution}. Our effectiveness on glow suppression and scalability in existing LLIE methods are validated on real-world datasets. Meanwhile, the proposed method does not require any pre-training.

\end{itemize}

\vspace{-7mm}
\section{Related Work}
In this section, we outline relevant works on deep learning-based methods for LLIE and light effects suppression approaches in fields such as nighttime dehazing.
\subsection{Low-light Image Enhancement}
The LLIE task aims to increase image visibility so as to benefit a series of downstream tasks (\textit{e.g.} classification, detection, and recognition). Benefiting from the flourishing of deep learning, a large proportion of existing LLIE methods provide outstanding results in typical low-light scenes. Part of these methods~\cite{zhang2019kindling,zhang2020self,zhang2021beyond} typically introduced the Retinex model~\cite{jobson1997multiscale} from traditional methods, enhancing the luminance and reflection components separately through dedicated sub-networks.
On another matter, the elimination of noise, an explicit term in the imaging system, is often decoupled into a separate sub-task~\cite{zhu2020zero,yang2021sparse,2106.14501,2111.15557}.

Furthermore, more recent attention has focused on realistic nighttime scenarios.
On the one hand, real night scenes lack paired datasets, hence EG~\cite{jiang2021enlightengan} and DCE~\cite{li2021learning} escaped the reliance on paired datasets by utilizing an unsupervised generative adversarial network and designing image-specific curves, respectively. On the other hand, some approaches~\cite{zhang2019zero,zhu2020eemefn} attempted to confront more challenging light conditions like extreme low-light \cite{chen2018learning,wei2020physics}. 

However, all these methods fail to address the extensive and blurred glow caused by nighttime light sources, which can be further exposed with direct enhancement, thus resulting in even more information loss.

\subsection{Glow Suppression}
The primary object of glow effects suppression is the glow from multi-scattered near-field light sources (distinct from atmospheric light). Due to the scarcity of glow suppression methods in LLIE, we conduct research in adjacent areas.

Generally, the glow formation function is referred to as APSF. Since glow causes a specific range of information loss, modeling APSF is essential to improve the performance of outdoor vision systems. 
\cite{narasimhan2003shedding} introduced APSF into computer vision for the first time and designed a physics-based model to describe multiple scattering.
As it is well suited for severe weather such as fog, haze and rain, \cite{li2015nighttime} brought the model into nighttime haze removal and utilized the short-tailed distribution of glow on the gradient histogram to separate the glow layer, followed by \cite{park2016nighttime,yang2018superpixel}.
Lately, in response to noise, flare, haze, and blur in Under-Display Camera (UDC) images, \cite{feng2021removing} defined a physics-based image formation model to analyze and solve the aforementioned degradations. Besides, the authors measured the real point spread function (PSF) of the UDC system and provided a model-based data synthesis pipeline to generate realistic degraded images. But the specific glow pattern in the UDC system is not applicable in a wide range of real scenes, and it is not feasible to measure true PSFs in each case.

In LLIE, \cite{sharma2021nighttime} is the first to focus on glow effects suppression. More recently, the same research team introduced an unsupervised method \cite{2207.10564} that integrates a layer decomposition network and a light-effects suppression network. Yet a rudimentary frequency-based decomposition struggles to effectively separate glows from irregularly shaped light sources or those with uneven intensity. This is due to the absence of interpretable physical model support.

In contrast, the proposed method is fully theoretically informed, as evidenced by a physical model designed on glow formation, thus allowing for the effective suppression of various shapes and intensities of glows shown in Fig. \ref{shape}. Moreover, unlike the above works requiring training datasets with and without glow, our approach does not rely on any paired or unpaired datasets, nor does it require pre-training.

\begin{figure}[t]
\centering
\includegraphics[width=0.49\textwidth]{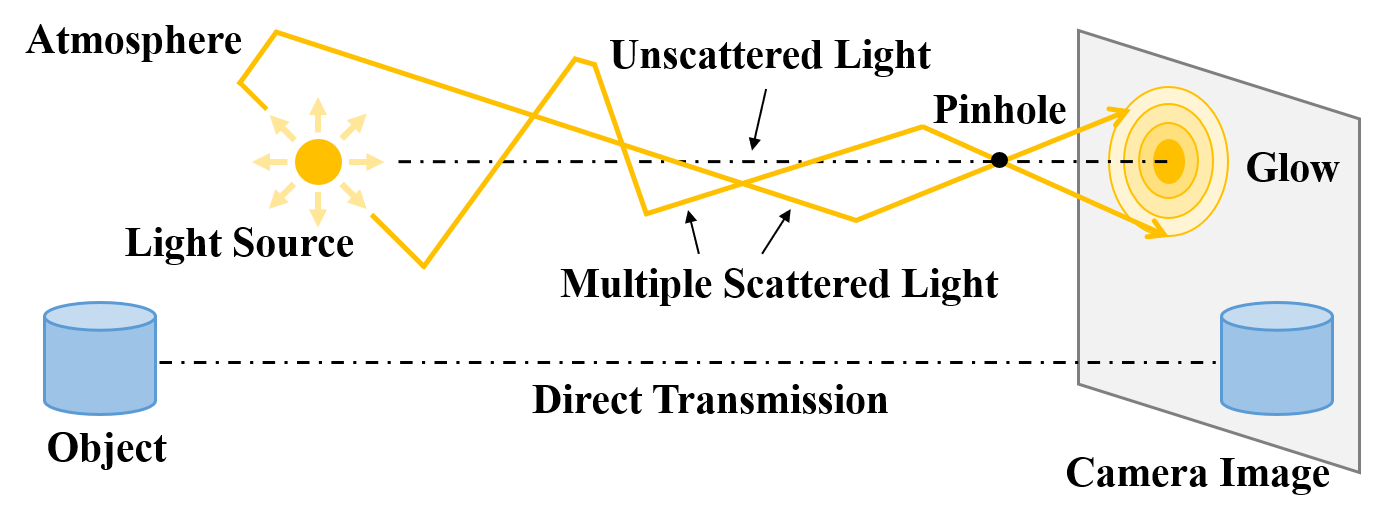} 
\caption{Glow formation. Light sources are scattered several times to reach the observer, forming blurred glows that vary in shape.
}
\label{multi}
\end{figure}

\begin{figure*}[t]
\centering
\includegraphics[width=0.96\textwidth]{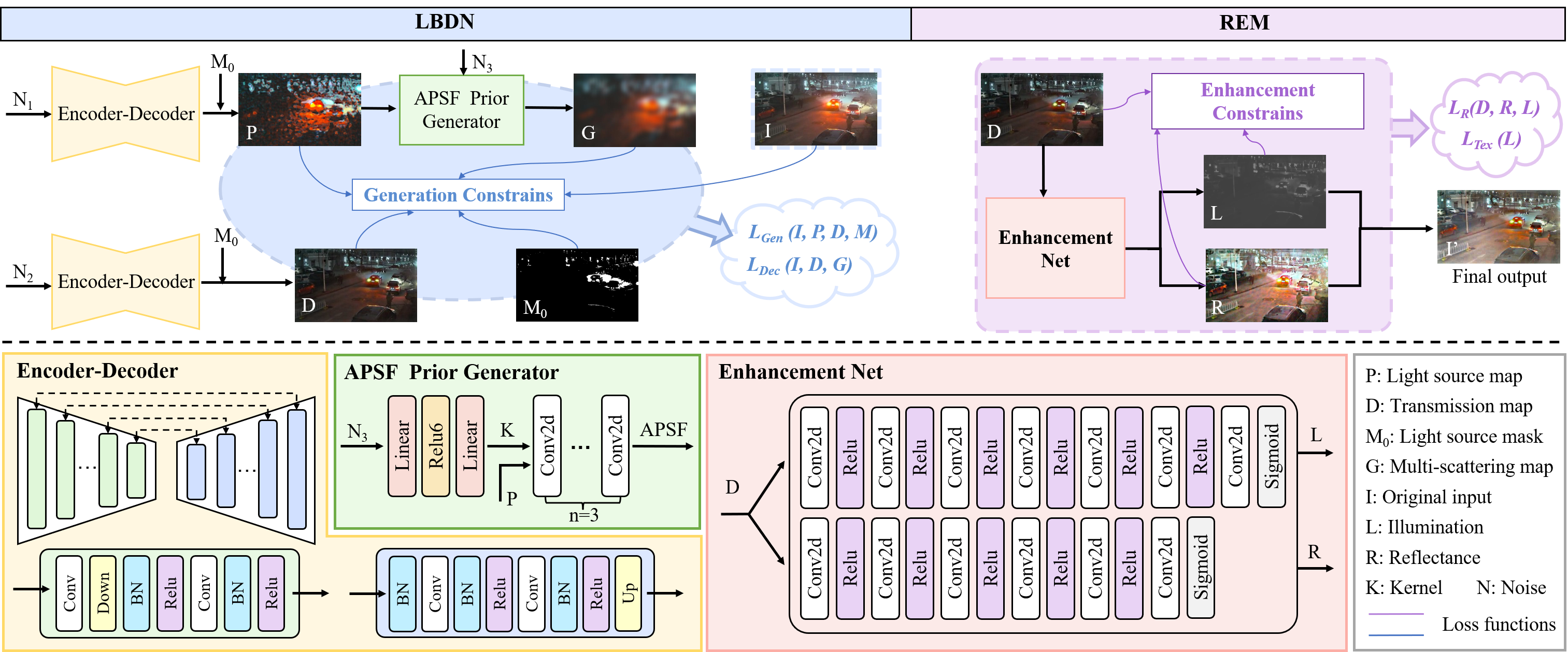}
\caption{Overall framework of the proposed method, consisting mainly of a glow suppression module LBDN and an enhancement module REM. With random noises as inputs, the LBDN separates the multiple scattering map $G$ and the direct transmission $D$ by encoder-decoder with the guidance of the priors $M$ and APSF. Then $D$ is subsequently fed into the REM for light correction.}
\label{framework}
\end{figure*}
\vspace{5mm}
\section{Physical Model}
\subsection{APSF-based Physical Glow Formation}
As the glow formation described in Fig.~\ref{multi}, rays from the light source are scattered several times by particles, and further interactions with different angles overlap on the image to form glow. Finally, the glow in the image plane forms a round region, which has a maximum intensity in the center point and decreases towards the surroundings.

To obtain the multi-scattering map of irregular light sources, the imaging of glow is derived by the intensity of the multiply scattered radiations $I(T, \mu)$  from a point source with APSF \cite{chandrasekhar2013radiative}, which is related to the forward scattering parameter $q$, the phase function $P(cos\alpha)$ of the particles in the atmosphere, and the optical thickness of the atmosphere $T$. 
The solution to $I(T, \mu)$ is given by expanding the Henyey-Greenstein phase function~\cite{ishimaru1978wave} in terms of Legendre polynomials. Please see the supplementary material for more detailed formulations.

The glow around one light source of arbitrary shape $S$ is formed as a combination of multiple isotropic light source elements with different radiation intensities $I_{0}(x, y)$. With light from each source element approximately passing through the same atmosphere and, based on the superimposability of light, the intensity $I_{0}$ of different light source elements can be summed to yield the overall intensity $I_{total}$ :
\begin{equation}
\small
\begin{array}{c}
I_{total} = \sum_{i} (I_{0} \otimes APSF_{i}) = (I_{0}\times S) \otimes A P S F,  i \in S.
\end{array}
\small
\end{equation}

Accordingly, the connection between the image $P$ with ideal light sources and the final imaging result $G$ after suffering from multiple scattering can be established as:
\begin{equation}
\begin{array}{c}
G= P \otimes A P S F.
\label{glowterm}
\end{array}
\end{equation}

In this paper, we take advantage of deep learning to estimate the APSF via a kernel estimation network to enable the injection of this physical prior into our network as guidance.

\subsection{Nighttime Imaging With Light Sources}
In this section, we refer to existing physical models from neighboring domains to design our nighttime imaging with near-field light sources. In bad weather, the light source contributes to two light components, one is the light reflected from the target through particle attenuation, and the other is the atmospheric light formed by the scattering of ambient light in the infinite distance, in correspondence to the direct transmission $D(x, \lambda)$ and atmospheric light $A(x, \lambda)$ in the atmospheric scattering model, with $x$ and $\lambda$ representing the position of the pixel in the image and the wavelength of light:
\begin{equation}
\begin{array}{c}
I(x, \lambda)=D(x, \lambda)+A(x, \lambda)\\=e^{-\hat{\beta}(\lambda) d(x)} R(x, \lambda)+L_{\infty}\left(1-e^{-\hat{\beta}(\lambda) d(x)}\right),
\label{haze}
\end{array}
\end{equation}
where $I(x, \lambda)$ is the degraded image captured by a camera and $R(x, \lambda)$ denotes the ideal image to restore; $L_{\infty}$ represents the value of atmospheric light at infinity; and $t=e^{-\hat{\beta}(\lambda) d(x)}$ represents the transfer function with the physical meaning referring to the proportion of light, which reaches the image plane via attenuation light through atmosphere particles.

While in the inference of our model upon Eq.~\ref{haze}, the following three items were sequentially reconsidered: 

\textbf{1) Atmospheric light:}  
Contrary to the atmospheric scattering model aiming at imaging bad weather, we serve for the LLIE task, whose target scene is a clear night.
As the nighttime atmospheric light value $L_{\infty}$ and the total scattering coefficient $\hat{\beta}(\lambda)$ in a clear night both converge to zero, the atmospheric light $A(x, \lambda)$ in our model can be safely ignored. 

\textbf{2) Multi-scattering:} Since $D(x, \lambda)$ is the direct transmission after removing the scattered flux of all incident energy, excluding the glow formed by scattering, which is of our concern. For that, we additionally consider the prevalence of active light sources at night and represent the result of their multiple scattering as $G(x, \lambda)$ addition to the model, with its modeling equivalent to Eq. \ref{glowterm}. 

\textbf{3) Direct transmission:} Inspired by Retinex theory~\cite{jobson1997multiscale}, our transmission map $D$ reaching the observer can be decomposed into an illumination map $L$ and an illumination-independent reflectance $R$. Such decomposition helps to subsequently correct the illumination conditions individually for better visibility.

Adapting Eq. \ref{haze} to the above derivation, our final nighttime imaging model NIM-NLS can be expressed as:
\begin{equation}
\begin{array}{c}
I(x, \lambda)\!=\!D(x, \lambda)\!+\!A(x, \lambda)+\!G(x, \lambda)\\
=\!D(x, \lambda)\! + 0 +\!G(x, \lambda)\\
=\!R\times L\! +\!P \otimes A P S F,
\label{model}
\end{array}
\end{equation}
where $P$ is the light source map indicating the positional information perceived from light sources in glow.

\begin{figure*}
\centering
\includegraphics[width=0.98\textwidth]{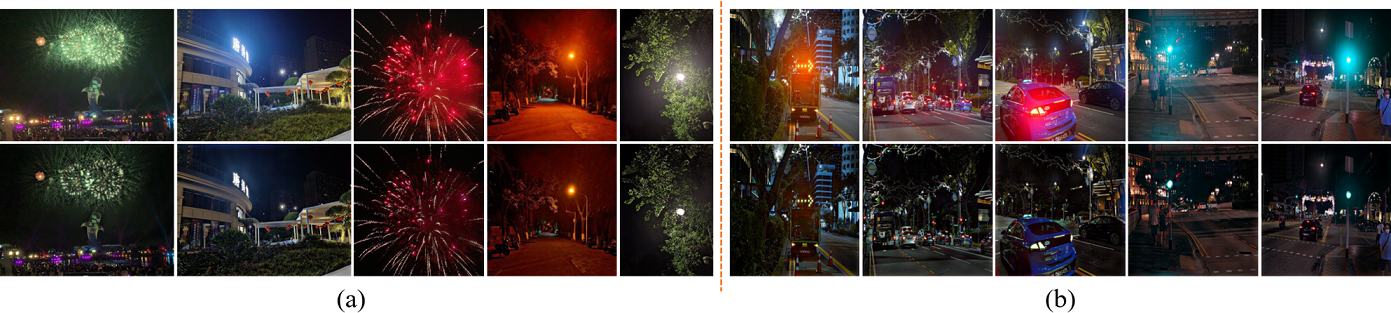}
\caption{Glow suppression results from our LBDN on (a) our self-collected images suffering from different types of glow effects, and (b) night-time traffic scenes on the light-effects dataset~\protect\cite{sharma2021nighttime}. Zoom in to see details.}
\label{ourdata}
\end{figure*}

Our physical imaging model, NIM-NLS, describes the actual generation of nighttime glows without any assumptions or restrictions on the intensity, and in which APSF is proven capable of modeling the scattering of arbitrarily shaped light sources, effectively addressing the two challenges of real-world glows (Fig.~\ref{ourdata}).
\vspace{3mm}
\section{Network}
To solve for NIM-NLS, we propose a network (Fig.~\ref{framework}) incorporating a Light-aware Blind Deconvolution Net (LBDN) and a Retinex-based Enhancement Module (REM), with the former aiming to obtain the glow-free $D$, and the latter to restore the low-light $D$ for an enhanced one.

\subsection{Light-aware Blind Deconvolution Net}
On the basis of the Maximum a Posterior (MAP), our objective can be reformulated for deep learning from Eq.~\ref{model} as:
\begin{equation}
\begin{array}{c}
(\mathbf{d},\! \mathbf{p}, \mathbf{apsf})\!=\!\mathop{\arg \min} \limits_{(\mathbf{d}, \mathbf{p}, \mathbf{apsf})}\!\|\mathbf{P}\! \otimes\! \mathbf{APSF}\! +\!\!\mathbf{D}\!\!-\!\mathbf{I}\|^{2}\\
+\lambda \phi(\mathbf{D})+\tau \varphi(\mathbf{P} ) + \omega \rho({\mathbf{APSF})}, \\

0\!\leq\!d_{i}\!\leq\!1,0\!\leq p_{j} \leq 1,apsf_{z}\!\geq\!0,\\

\!\sum_{z}\! apsf_{z}\!=\!1,\forall i,j,z,
\label{target}
\end{array}
\end{equation}
where $\|\mathbf{P} \otimes \mathbf{APSF} +\mathbf{D}-\mathbf{I}\|^{2}$ is the fidelity term, and  $\phi(\mathbf{D})$, $\varphi(\mathbf{P})$, $\rho(\mathbf{APSF})$ are three regularization terms, corresponding to constraint the generation $D$, $P$, and the APSF, with the relevant functions specified in Loss Function. $\lambda$, $\tau$, and $\omega$ are regularization parameters.

The task of estimating the blur kernel and solving for the sharp image is commonly known as blind deconvolution~\cite{shan2008high,almeida2009blind}. 
Unlike existing global blind deconvolution models, our model should first decompose the light source $P$ locally present in $I$ and then simulate the glow with the APSF obtained from the kernel estimation network. Hence, we design a Light-aware Blind Deconvolution Net (LBDN) to reach achieve~Eq.~\ref{target}, motivated by Double-DIP~\cite{gandelsman2019double} decomposing of a complex image into two or more simple layers. In this case, two DIP generators ~\cite{ulyanov2018deep} are applied for the light source map $P$ and the direct transmission map $D$ from random noise. Besides, we additionally introduce two priors. One is a mask $M_{0}$ that marks the shape and location of the light source. The mask derived by setting an illumination threshold of a certain intensity constrains the generation of $D$ and $P$. The other prior is the APSF, acquired via iterative convolution $n$ times with the blur kernel $k$ from the blur kernel estimation net. It guides the simulation of the multi-scattering map $G$. Since DIP is suitable for capturing statistical information from natural images while limited in estimating blur kernel priors \cite{ren2020neural}, we generate the kernel via a simpler full-connected net (FCN).

\subsection{Retinex-based Enhancement Module} 
The Retinex model has been proven valid throughout extensive previous LLIE works \cite{zhu2020zero,zhang2020self,liu2021retinex,wu2022uretinex}.
In the Retinex-based Enhancement Module (REM), we decompose the direct transmission map $D$ reaching the observer into reflection $R$ and illumination $L$ according to Retinex theory. $R$ is always constant under different illumination conditions, so we correct the illumination by gamma transformation for better visibility. In particular, our REM and LBDN are co-related in that they maintain a consistent zero-shot learning approach without any pre-training, which is achieved through a series of non-reference loss functions.

\subsection{Loss Function}
Our total loss function consists of 
the generation constraint and the enhancement constraint. 
The former guides the generation of individual layers in the decomposition, and the latter ensures the illumination recovery of the final output.

\paragraph{Generation constraints.} 
Decomposition loss $L_{Dec}$ measures the absolute difference between the recomposition of the different layers obtained from Eq.~\ref{model} and the input image by constraining the sum of the decomposition layers to fit all the information in the original image as closely as possible.

\begin{equation}
\begin{array}{c}
 L_{Dec}(I',I) = ||D + G - I||_{1}.
\end{array}
\end{equation}

In addition, we use the generation loss $L_{Gen}$~\cite{gandelsman2019double} to guide the construction of $P$ and $D$, with the light source spatial location mask $M_{0}$ serving as prior information for the location of light sources.

\paragraph{Enhancement constrains.} 
In REM, the Retinex loss $L_{R}$ shows the absolute difference between the recomposed $D'$ from REM and the output $D$ from LBDN. Since the maximum channel of the reflectance coincides with that of the original image, we add this constraint to separate the reflectance.

\begin{small}
\begin{equation}
\begin{array}{c}
 L_{R}(R,\!L,\!D)\!\!=\!\!||R\!\times\!\!L\!\!-\!\!D||_{1}\!\!+\!||F_{c}(D,x)\!\!-\!\!F_{c}(R,x)||_{1},\\

F_{c}(I,x) = \mathop {\max }\limits_{c \in \{ R,G,B\} } {\rm{ }}{I^c}({\rm{x}}).
\end{array}
\end{equation}
\end{small}

To ensure that the texture of $R$ is clear, we also introduce the texture enhancement loss $L_{Tex}$ in \cite{zhu2020zero} to strengthen the smoothness of $L$.

We multiply each non-reference loss with its respective weight, where the weights of generation constraints are $1.0$, while enhancement losses' are set to $0.5$.
\begin{center} 
\begin{figure*}[!t]
	\centering
    \footnotesize{
    \tabcolsep=1pt
      \begin{tabular}{cccccccc}
      \includegraphics[width=0.12\textwidth]{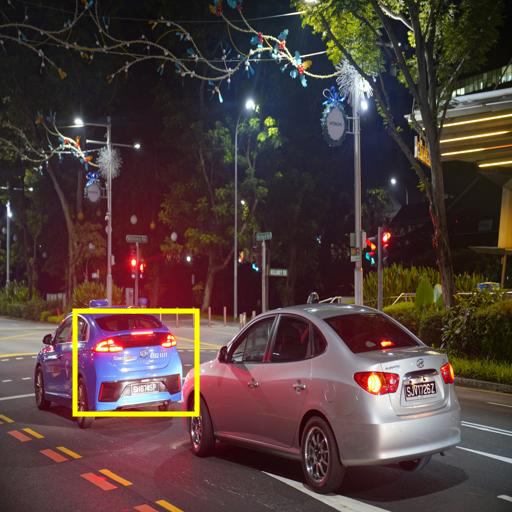} &
      \includegraphics[width=0.12\textwidth]{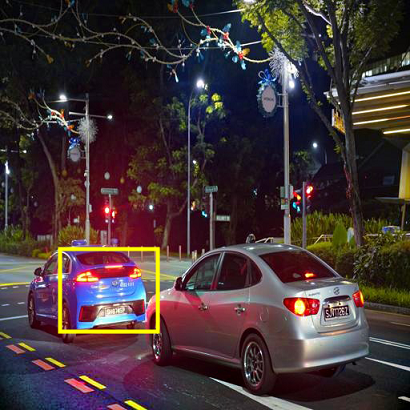} &
      \includegraphics[width=0.12\textwidth]{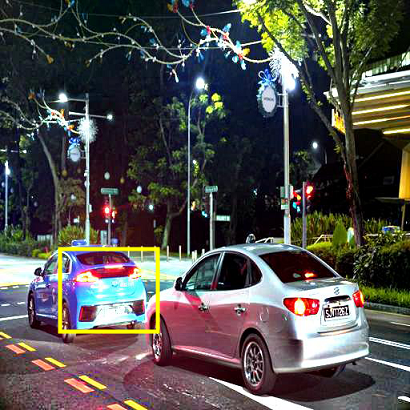} &
      \includegraphics[width=0.12\textwidth]{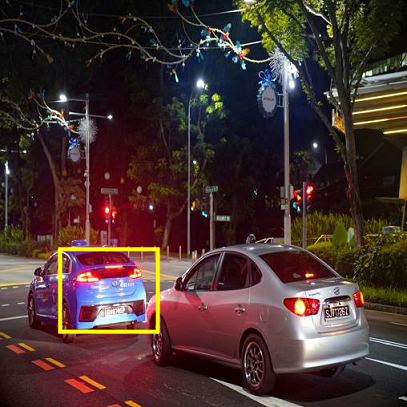} &
      \includegraphics[width=0.12\textwidth]{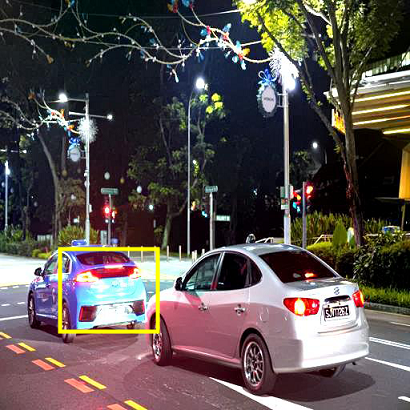}&
      \includegraphics[width=0.12\textwidth]{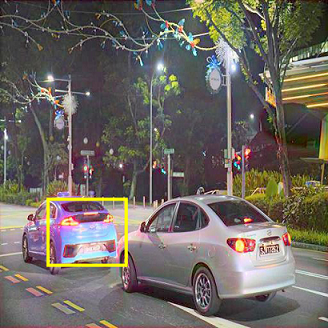}&
      \includegraphics[width=0.12\textwidth]{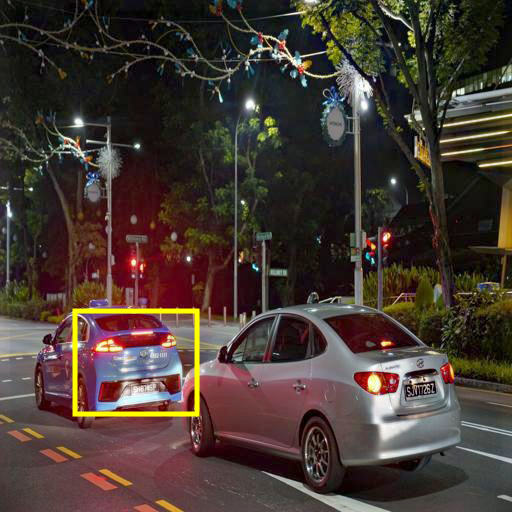}&
      \includegraphics[width=0.12\textwidth]{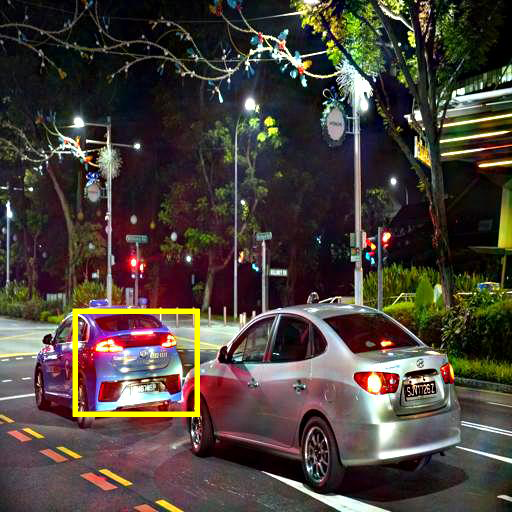} \\
      
      \includegraphics[width=0.12\textwidth]{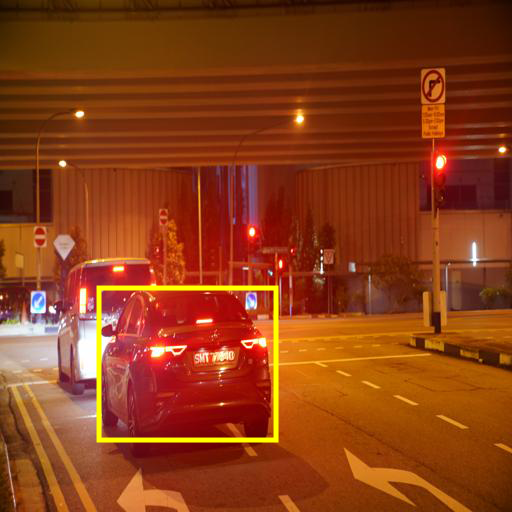} &
      \includegraphics[width=0.12\textwidth]{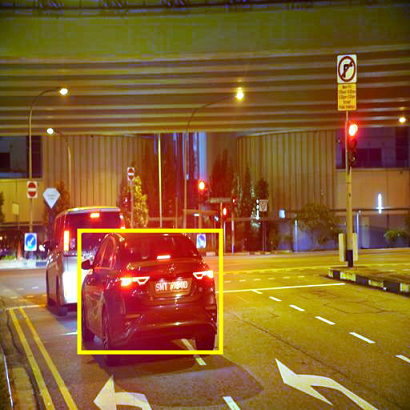} &
      \includegraphics[width=0.12\textwidth]{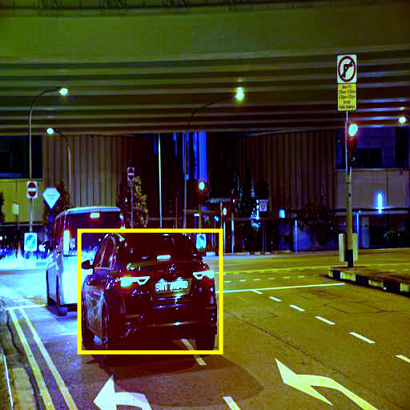} &
      \includegraphics[width=0.12\textwidth]{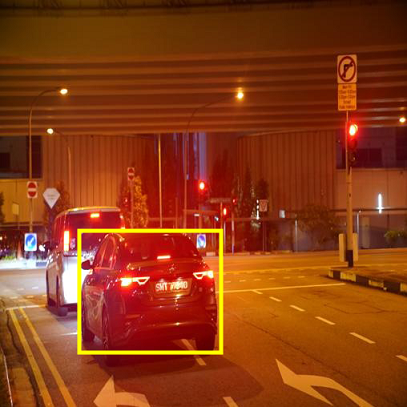} &
      \includegraphics[width=0.12\textwidth]{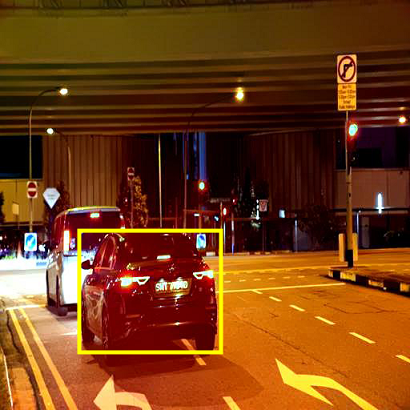}&
      \includegraphics[width=0.12\textwidth]{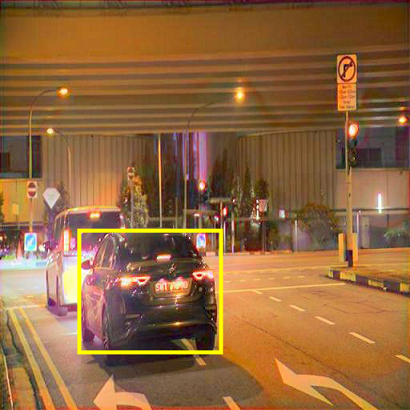}&
      \includegraphics[width=0.12\textwidth]{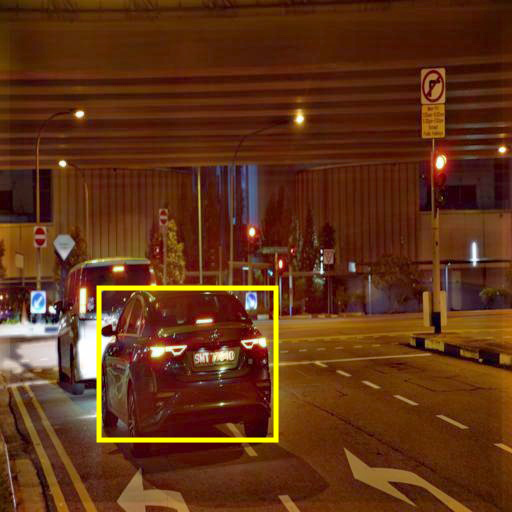}&
      \includegraphics[width=0.12\textwidth]{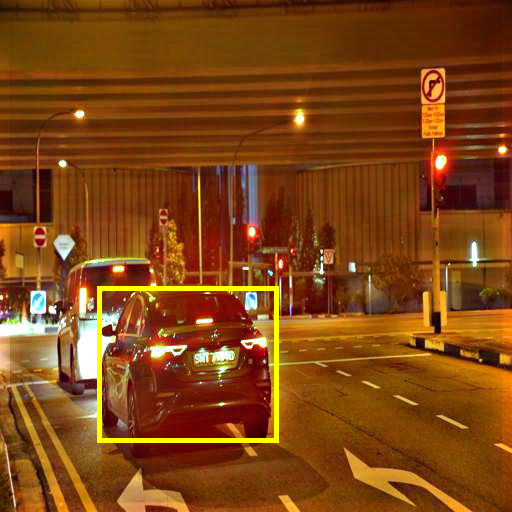} \\
      
      \includegraphics[width=0.12\textwidth]{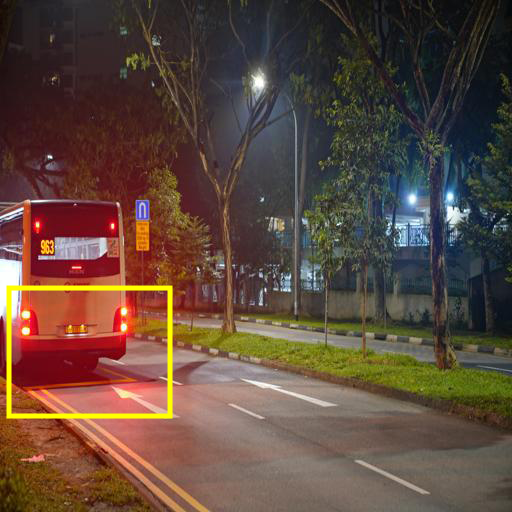} &
      \includegraphics[width=0.12\textwidth]{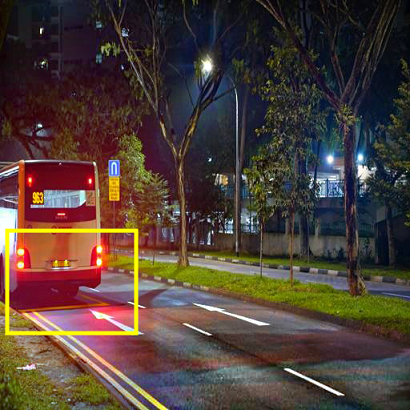} &
      \includegraphics[width=0.12\textwidth]{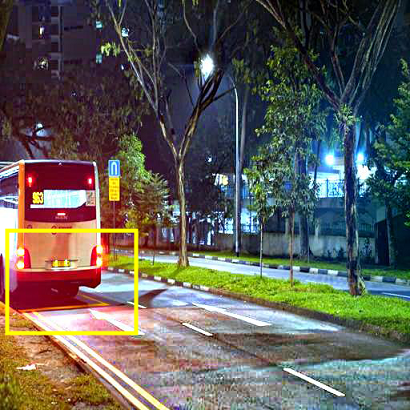} &
      \includegraphics[width=0.12\textwidth]{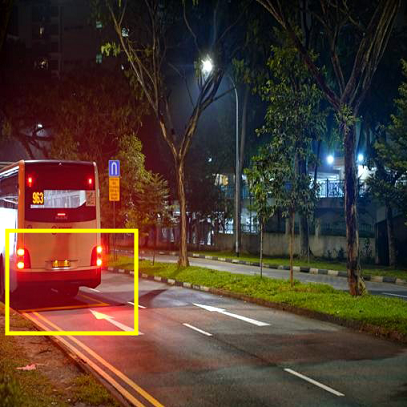} &
      \includegraphics[width=0.12\textwidth]{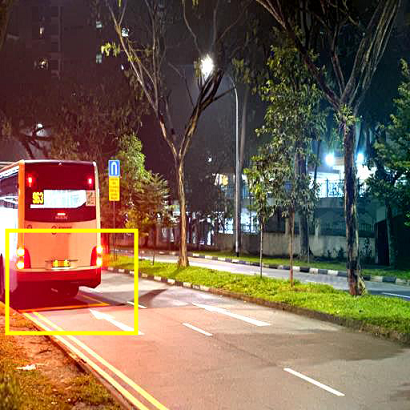}&
      \includegraphics[width=0.12\textwidth]{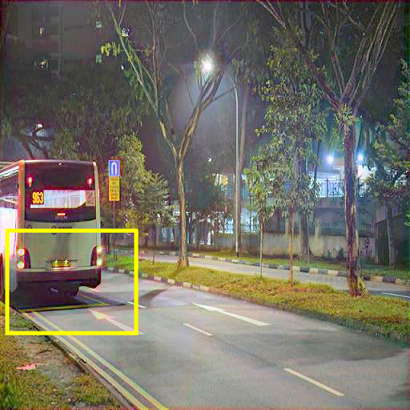}&
      \includegraphics[width=0.12\textwidth]{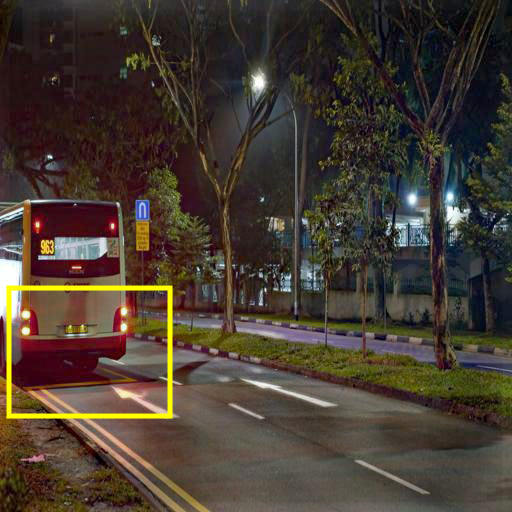}&
      \includegraphics[width=0.12\textwidth]{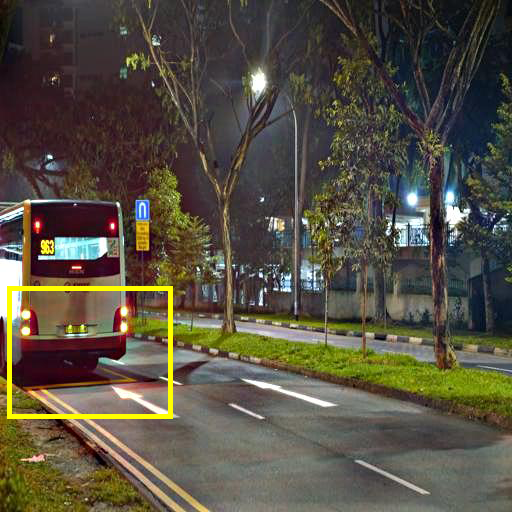}\\
      \small{Input}& \small{3R}& 
      \small{NHR}&
      \small{DCP}&
      \small{UNIE}&
      \small{Sharma}& 
      \small{Our LBDN}& \small{Our REM}\\
      \end{tabular}}
	\caption{Some examples of visual comparisons on the light-effects dataset~\protect\cite{sharma2021nighttime}. Our LBDN achieves the most visually significant glow suppression, and REM further reinforces the visibility with appropriate luminance adjustments.}
 \label{comparison}
\end{figure*}
\end{center}
\begin{center} 
\begin{figure}[t]
	\centering
    \includegraphics[width=0.44\textwidth]{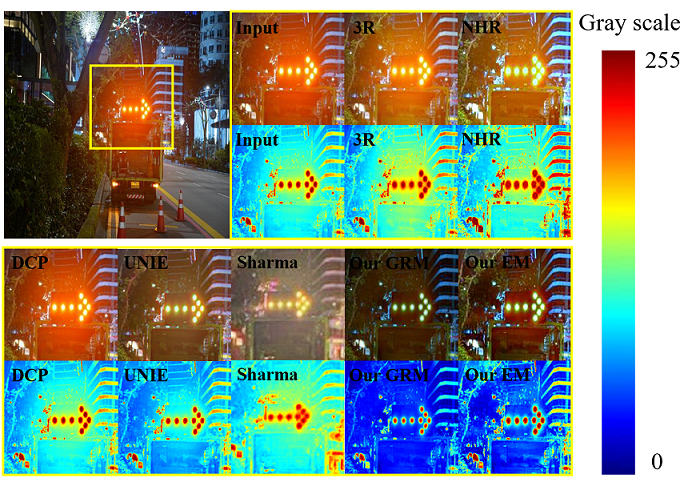} 
	\caption{Some examples suffering from the glow and their corresponding heat maps. Dark areas and distinct color borders of the glow in the heat map show non-smooth intensity variations.}
\label{glowPatch}
\end{figure}
\end{center}
\vspace{-6mm}
\begin{figure*}
\centering
\includegraphics[width=0.99\textwidth]{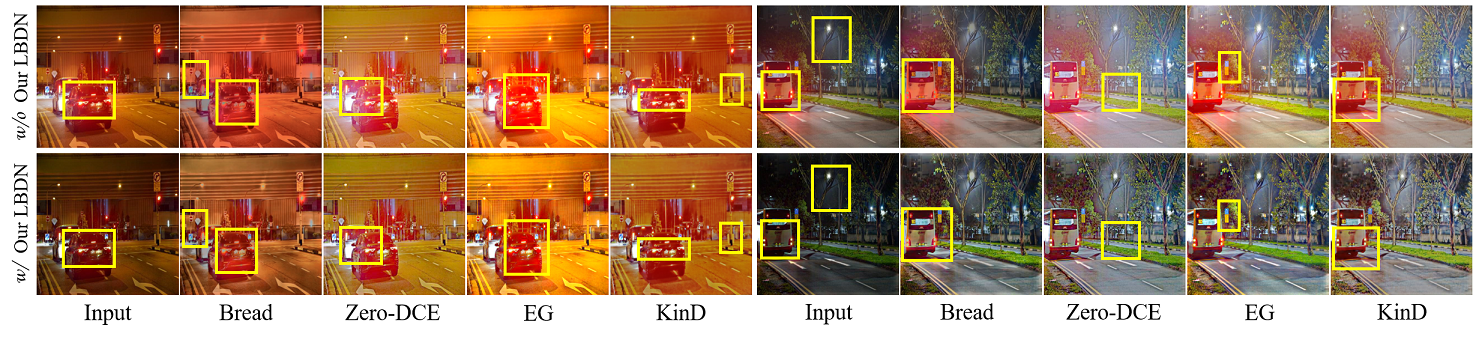}
\caption{Two groups of comparison between enhancement results $w/o$ and $w/$ glow suppression. The first row of each set shows the nighttime image affected by glows and the enhanced outputs with SOTA methods in LLIE. And the second row presents the glow-free image from the proposed LBDN and the corresponding later enhanced results.}
\label{LLIE}
\end{figure*}
\vspace{-3mm}
\section{Experiment}
\subsection{Experimental Settings}
\paragraph{Implementation details.} 
Our implementation is done with PyTorch on an NVIDIA GeForce RTX 3060 GPU. Since our model is based on zero-shot learning, which trains and tests each individual input to obtain the specific glow-free result, pre-training is not required.

\paragraph{Datasets.} 
We test on two groups of real-world datasets. The light-effects dataset~\cite{sharma2021nighttime} contains 500 images with multiple light colors in various scenes. Besides, we take 30 images with glow in different shapes and uniform intensity with Huawei Note 20 and Redmi Note 10 Pro.

\subsection{Comparison on Glow Suppression}
We show the effective suppression of a wide variety of glows in Fig.~\ref{ourdata}. Our method performs excellent suppression when confronted with glows from different scenes, such as streetlights and headlights with a very wide luminous area, fireworks with dense and unevenly distributed light points, and billboards with irregular shapes.

To further show the superiority of the proposed method in improving nighttime visibility, we compared it with two existing glow suppression methods, UNIE \cite{2207.10564} and Sharma\cite{sharma2021nighttime}. 
As there is minimal work on glow suppression, supplementary consider three nighttime dehazing methods, including 3R \cite{zhang2020nighttime}, NHR \cite{li2015nighttime} and DCP \cite{he2010single}.
The subjective results are in Fig.~\ref{comparison}. 
Our method provides effective suppression in regions of uneven intensity or high frequencies in the glow, as evidenced by the zoomed-in local glow zone in Fig.~\ref{glowPatch} together with the heat maps (more visual results can be seen in the supplementary material).
In Table.~\ref{comparisonTable}, two metrics from dehazing are borrowed to test the clarity of glow patches in each image. The metrics e and r \cite{hautiere2008blind} indicate, respectively, the ratio of new visible edges and the percentage of saturated pixels produced. 
For the reconstruction quality of the whole image, 
we believe a user study is highly recommended, given the lack of recognized objective metrics for glow suppression. Therefore, 20 participants were involved in ranking these methods based on 1) glow suppression effectiveness, 2) visibility, and 3) the presence of color bias and artifacts, with 1 to 7 in order of worst and best.
\vspace{-1mm}
\begin{center} 
\begin{table}[t]
\renewcommand{\arraystretch}{1.3}
\centering
\scalebox{0.74}{
\setlength{\tabcolsep}{1.0mm}{
    \begin{tabular}{c|c|cc|cc}
    \hline
    \multicolumn{2}{c|}{\multirow{2}{*}{Method}}&\multicolumn{2}{c|}{Local Glow}&\multicolumn{1}{c}{Global Image} \\ 
     \cline{3-5} 
    \multicolumn{2}{c|}{}
    & e $\uparrow$ & r $\uparrow$ 
    & User Study $\uparrow$ \\ 
    \hline
    \multirow{3}{*}{ND} 
    & 3R~\cite{zhang2020nighttime} 
    & 0.0632 & 1.2121 
    &3.4386\\
    & NHR~\cite{li2015nighttime} 
    & 0.1148 & \textcolor{red}{1.7988} 
    &3.3810\\
    & DCP~\cite{he2010single} 
    & 0.0313 & 1.0567 
    &2.8095\\
    \hline
    \multirow{4}{*}{GR} 
    & UNIE~\cite{2207.10564} 
    & 0.0520 & 1.5400 
    &3.8572\\  
    & Sharma~\cite{sharma2021nighttime} 
    & 0.0024 & 1.4876 
    &2.5238\\  
    & Our LBDN 
    & \textcolor{blue}{0.1516} & 1.0071 
    & \textcolor{blue}{5.9048}\\
    & Our REM 
    & \textcolor{red}{0.1743} &\textcolor{blue}{1.7540} 
    &\textcolor{red}{6.0952}\\
    \hline
    \end{tabular}}}
    \caption{Objective evaluation of representative nighttime dehazing methods (ND) and all existing glow removal methods (GR) on the light-effects dataset~\protect\cite{sharma2021nighttime}. \textcolor{red}{Red} and \textcolor{blue}{blue} are the best and second best results respectively.}
\label{comparisonTable}
\end{table}
\end{center}  
\vspace{-5mm}
\subsection{Low-light Image Enhancement}
Our initial aim in suppressing glow effects is to address the problem of further spread of glow in LLIE resulting in sharpness degradation or even color cast and artifacts. To validate our scalability, we combine the existing SOTA methods in LLIE to compare the performance before and after using our LBDN. The chosen methods includes KinD~\cite{zhang2019kindling}, 
 Zero-DCE~\cite{li2021learning}, EnlightenGan~\cite{jiang2021enlightengan} and Bread~\cite{2111.15557}.
The subjective and objective results on the light-effects dataset~\cite{sharma2021nighttime} are presented in Fig.~\ref{LLIE} and Table.~\ref{LLIETable} respectively, indicating that our LBDN significantly moderates the loss of information caused by further diffusion of glow during enhancement, as can be seen from the license plate in the first set of images and the ground in the second set. More display results can be found in the supplementary material.
\begin{center} 
\begin{figure}
	\centering
    \footnotesize{
    \tabcolsep=1pt
      \begin{tabular}{ccccc}
      \includegraphics[width=0.092\textwidth]{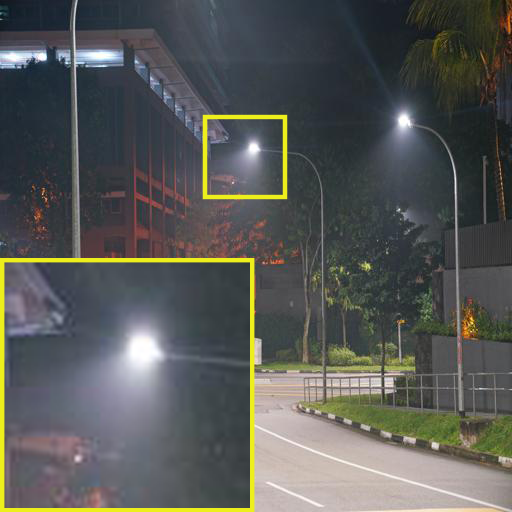} &
      \includegraphics[width=0.092\textwidth]{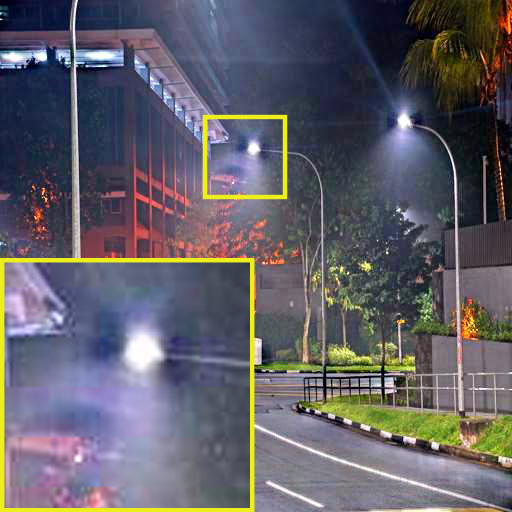} &
      \includegraphics[width=0.092\textwidth]{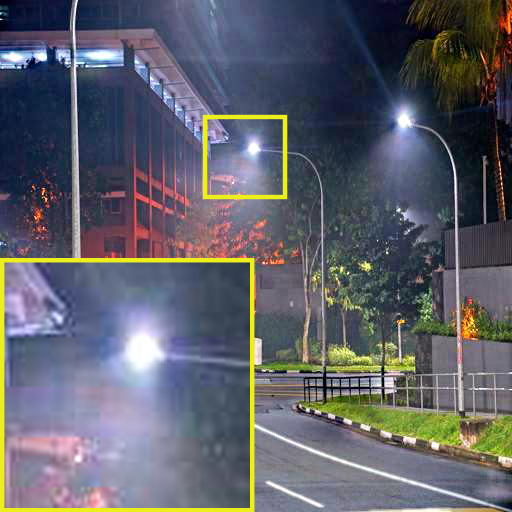} &
      \includegraphics[width=0.092\textwidth]{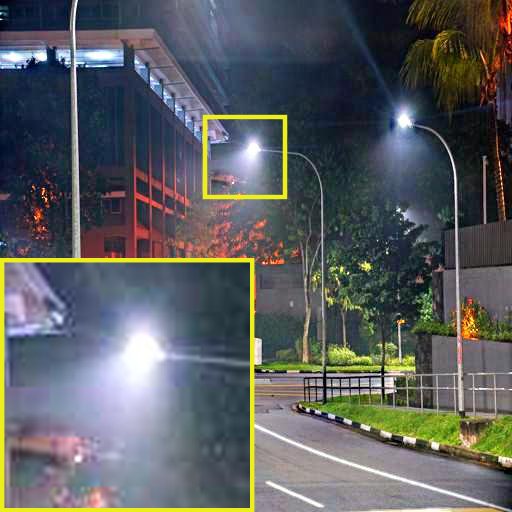}&
      \includegraphics[width=0.092\textwidth]{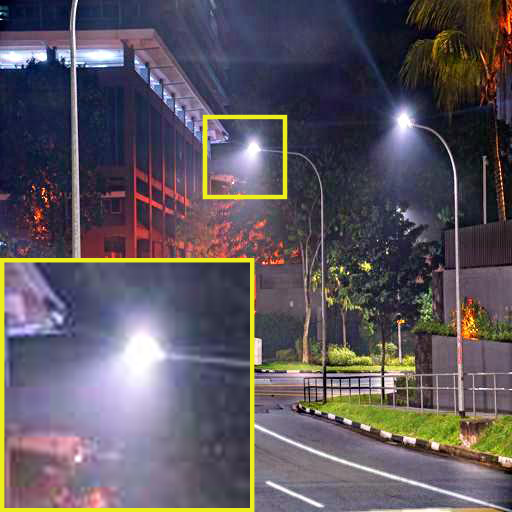}\\
      \footnotesize{Input}& \footnotesize{$n\!=\!1$}& \footnotesize{$n\!=\!2$}& \footnotesize{$n\!=\!4$}& \footnotesize{$n\!=\!5$}\\

      \includegraphics[width=0.092\textwidth]{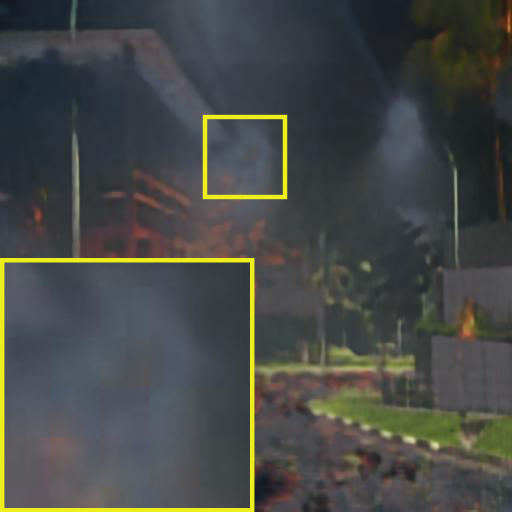} &
      \includegraphics[width=0.092\textwidth]{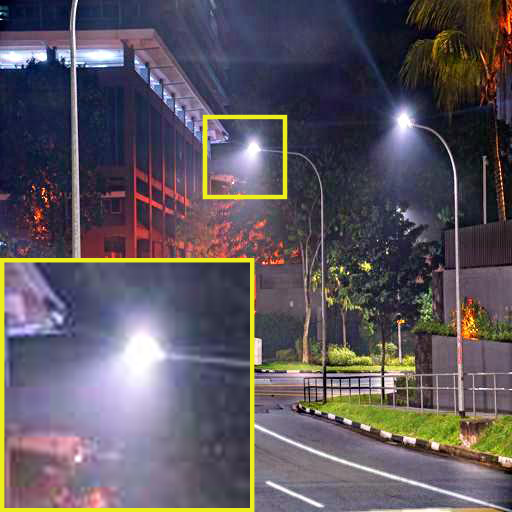} &
      \includegraphics[width=0.092\textwidth]{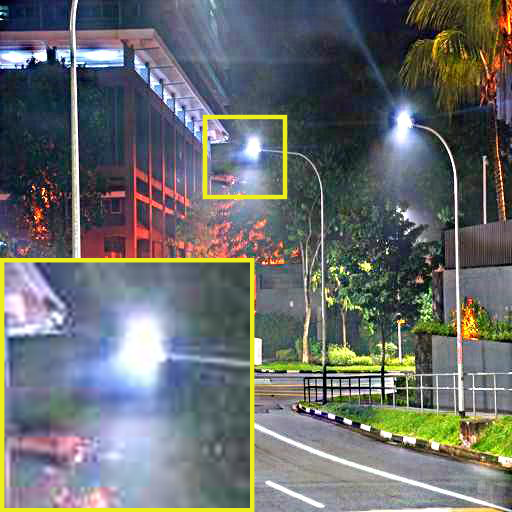} &
      \includegraphics[width=0.092\textwidth]{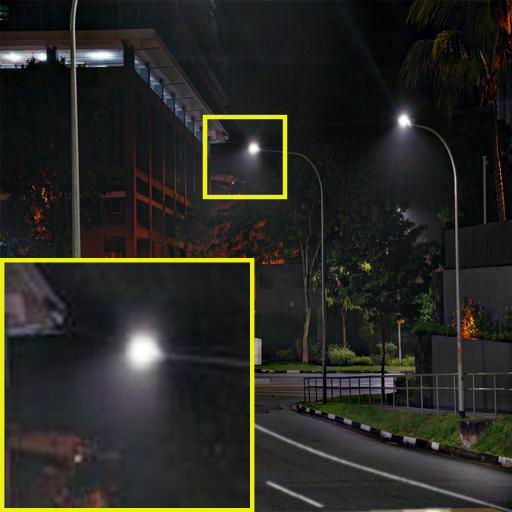} &
      \includegraphics[width=0.092\textwidth]{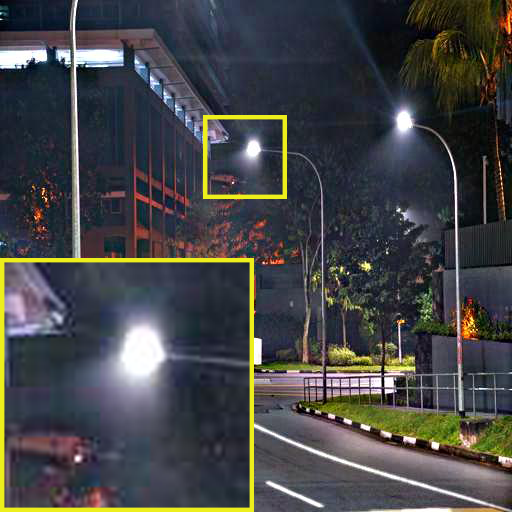}\\
      \footnotesize{$w/o$ APG}& \footnotesize{$w/o$ LMP}& \footnotesize{$w/o$  $L_{Tex}$}& \footnotesize{$w/o$  $F_{c}$}& \footnotesize{Ours}\\
      \end{tabular}}
	\caption{Ablation results of the light mask prior LMP, APSF prior generator APG and convolution number $n$ to simulate glow. In our model, $n$ is set to 3.}
	\label{ablation} 
\end{figure}
\end{center}
\vspace{-9mm}
\subsection{Ablation Study}
In this section, we demonstrate the effectiveness of some modules and parameters of the proposed method through ablation experiments. The subjective results and corresponding objective indicators are shown in Fig.~\ref{ablation} and Table ~\ref{AblationTable}.

\paragraph{Iterative convolutions for APSF.} To fit the APSF, we approximated the optimal solution by iterative convolutions, considering that the size of the convolution kernel limits the range of the simulated glow, whereby the number of iterations is represented by the parameter $n$. Although the objective index is not optimal when n is taken as $n=3$, we believe that the glow suppression and overall visibility improvement are better when combined with the visual outcome.

\paragraph{APSF prior generator (APG).} 
Our APSF prior generator serves as the core for glow suppression with direct decisions on the generation of $D$ and $G$. When removing the APM, we degrade the model to the basic Double-dip, which targets the light sources instead of irregular glow and offers almost no effect on glow suppression.

\paragraph{Light mask prior (LMP).} 
Since glows are present locally in the image, we perform light source segmentation by a light mask prior (LMP) before blind deconvolution. As shown in Fig.~\ref{ablation}, without LMP, the light source cannot be located accurately and thus the removal of the glow is very limited.

\paragraph{Loss function.} 
The key parameter $M_{0}$ (LMP) in generation losses has already been discussed, and the importance of composition losses is apparent since their removal will lead to the lack of principle for decomposition. Hence, we only focus on the ablation of $L_{Tex}$ and $F_{c}(I,x)$ here. Removing $L_{Tex}$ makes the decomposed illumination map patchy, resulting in unnatural artifacts in the output.
While the absence of $F_{c}(I,x)$ invalidates the decomposition of the illumination map for proper light correction.

\begin{table}[t]
\renewcommand{\arraystretch}{1.4}
\centering
\scalebox{0.7}{
\setlength{\tabcolsep}{1.0mm}{
\begin{tabular}{c|cccc|cccc}
\hline
\multirow{2}{*}{Method} & 
\multicolumn{4}{c}{Direct LLIE} & 
\multicolumn{4}{c}{Combined with our LBDN} \\
\cline{2-9}
                        & Bread  & Enlighten  & DCE  & KinD & Bread  & Enlighten  & DCE  & KinD\\ \hline
CEIQ $\uparrow$& 3.0434  & 3.3059  & 3.0861 &3.0162 & 3.2039  & 3.4003  & 3.3102 & 3.2940  \\
e $\uparrow$& 0.0836  & -0.1259  & -0.1244 &-0.1479 & 0.2304  & 0.1976  & 0.1753 & 0.1332  \\
r $\uparrow$& 1.5709  & 1.5213  & 1.5258 &1.5511 & 2.1696  & 2.3502  & 2.6556 & 2.5241\\ \hline
\end{tabular}}}
\caption{Objective evaluation on the impact of our LBDN on LLIE performance. With our LBDN as a pre-module to eliminate glow, all indicators have improved.}
\label{LLIETable}
\end{table} 
\vspace{-6mm}
\begin{table}[t]
\renewcommand{\arraystretch}{1.4}
\centering
\scalebox{0.7}{
\setlength{\tabcolsep}{1.0mm}{
\begin{tabular}{c|cccc|cccc|c}
\hline
\multirow{2}{*}{Module} & 
\multicolumn{4}{c|}{The value of $n$} & 
\multicolumn{4}{c|}{$w/o$} &
\multirow{2}{*}{Ours} \\
\cline{2-9}
                        & $n=1$  & $n=2$  & $n=4$  & $n=5$ & APG  & LMP  & $L_{Tex}$  & $F_{c}$&\\ \hline
e $\uparrow$& 0.3486  & 0.3523  & 0.3872 &0.4324 & -0.7115  & 0.2461  & 0.4652 & 0.4456 &  0.4334 \\
r $\uparrow$& 2.1281  & 2.0504  & 2.1712 &2.1954 & 1.0247  & 1.8514  & 3.1869 & 1.0586 &  2.0562 \\ \hline
\end{tabular}}}
\caption{The scores of the ablation results on the objective metrics e and r
, with higher scores indicating better visibility.}
\label{AblationTable}
\end{table} 
\vspace{3mm}
\section{Conclusion}
To remove irregular glows for nighttime visibility enhancement, we propose a novel approach for the first time viewing the glow suppression task as the learning of glow generation. We formulate a nighttime imaging model with near-field light sources, NIM-NLS, in which the APSF was first introduced into deep learning to learn the glow term. To fit this physical model, 
we accordingly design a light-aware blind deconvolution model LBDN and a subsequent Retinex-based enhancement network REM. The multi-scattering map can be estimated with the light source spatial location mask and the learned APSF as priors. Later in REM, the illumination of the clear transmission map is adjusted for better visibility. The proposed method does not rely on any paired or unpaired data, even without pre-training, and we demonstrated the method's effectiveness in extensive experiments.

\section*{Acknowledgments}
This work was supported by the Natural Science Foundation of China (62202347) and the Natural Science Foundation of Hubei Province (2022CFB578).

\bibliographystyle{named}
\bibliography{ijcai23}

\end{document}